\pdfoutput=1
\documentclass[runningheads]{llncs}
\usepackage{graphicx}
\usepackage{comment}
\usepackage{amsmath,amssymb} 
\usepackage{color}
\usepackage{float}
\usepackage{subfigure}
\usepackage{booktabs}
\usepackage{bm}
\usepackage{float}
\usepackage{url}
\usepackage{hyperref}
\usepackage{lipsum}

\usepackage[width=122mm,left=12mm,paperwidth=146mm,height=193mm,top=12mm,paperheight=217mm]{geometry}

\makeatletter
\newcommand{\@chapapp}{\relax}%
\makeatother
\usepackage[toc,page]{appendix}

\begin{document}
\pagestyle{headings}
\mainmatter

\title{Motion Guided 3D Pose Estimation from Videos} 



\titlerunning{Motion}

\author{Jingbo Wang\inst{1}\and
Sijie Yan\inst{1} \and Yuanjun Xiong\inst{2} \and Dahua Lin\inst{1}}
\authorrunning{Jingbo Wang et al.}
%
\institute{Department of Information Engineering, The Chinese University of Hong Kong
\email{\{jbwang,ys016,dhlin\}@ie.cuhk.edu.hk}\\
\and AWS/Amazon AI \email{yuanjx@amazon.com}}

\maketitle

\begin{abstract}
    We propose a new loss function, called motion loss, for the problem of monocular 3D Human pose estimation from 2D pose.
    In computing motion loss, a simple yet effective representation for keypoint motion, called pairwise motion encoding, is introduced.
    We design a new graph convolutional network architecture, U-shaped GCN (UGCN).
    It captures both short-term and long-term motion information to fully leverage the additional supervision from the motion loss.
    We experiment training UGCN with the motion loss on two large scale benchmarks: Human3.6M and MPI-INF-3DHP.
    Our model surpasses other state-of-the-art models by a large margin.
    It also demonstrates strong capacity in producing smooth 3D sequences and recovering keypoint motion.
    \keywords{3D Pose Estimation, Motion Loss, Graph Convolution}
\end{abstract}

\section{Introduction}


3D human pose estimation aims at reconstructing 3D body keypoints from thier 2D projections,
such as images~\cite{tome2017lifting,li20143d,tekin2016structured,pavlakos2017coarse},
videos~\cite{cheng2019occlusion,tekin2016direct},
2D pose~\cite{martinez2017simple,pavllo20193d,lin2019trajectory},
or their combination~\cite{park20163d,tekin2017learning}.
Unlike the 2D pose estimation,
this problem is ill-posed in the sense that the lack of depth information in the 2D projections input leads to ambiguities.
To obtain the perception of depth,
recent works~\cite{iskakov2019learnable,qiu2019cross} utilized multiple synchronized cameras for observing objects from different angles
and has achieved considerable progress.
However, compared with monocular methods, multi-view methods are not practical in reality because of their strict prerequisites for devices and environments.

Recent years, video-based 3D human pose estimation~\cite{cai2019exploiting,lin2019trajectory,lin2017recurrent,dabral2018learning} receives attention quickly.
Taking a video as input, models are able to perceive the 3D structure of an object in motion and better infer the depth information for each moment.
It significantly promotes the estimation performance under the monocular camera.
Unlike image-based models, video-based models~\cite{lin2019trajectory,cai2019exploiting} are supervised by a long sequence of 3D pose,
which increase the dimensionality of solution space by hundreds of times.
In most existing works, the common loss function for supervising 3D pose estimation models is \emph{Minkowski Distance}, such as $\ell_1$-loss and $\ell_2$-loss.
It independently computes the overall location error of the predicted keypoints in 3D space with respect to their ground-truth locations.

\begin{figure*}[t]
    \centering
    \includegraphics[width=1.0\linewidth]{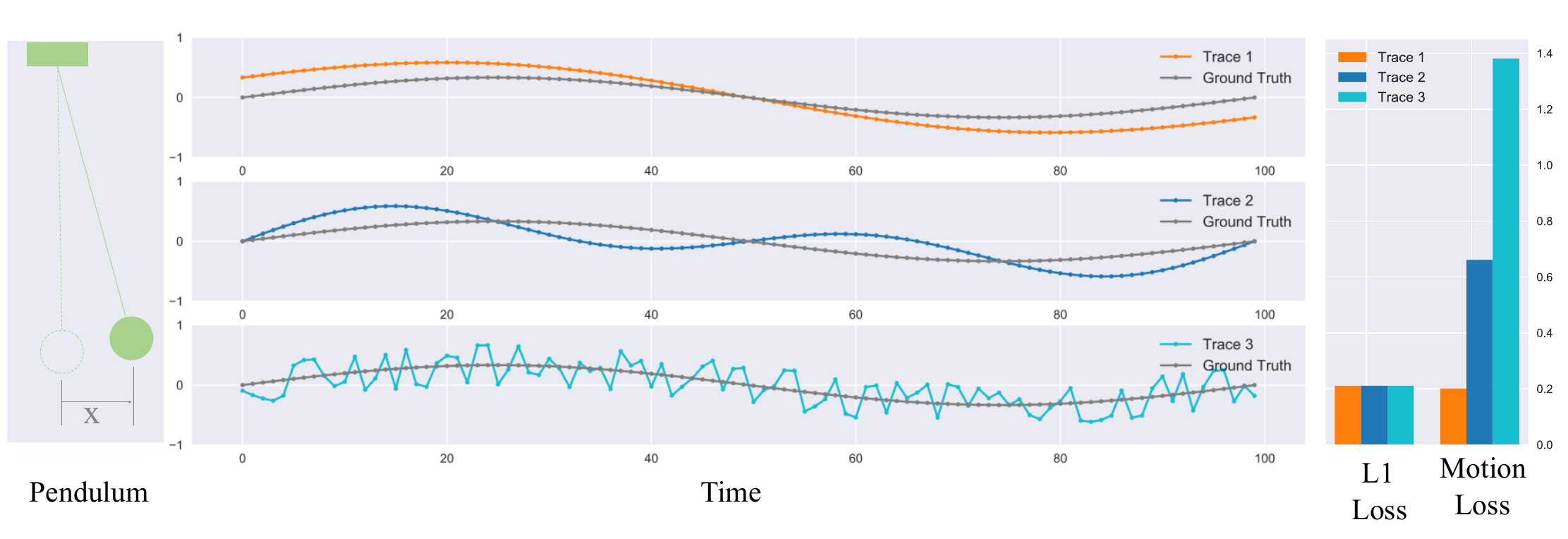}
    \caption{\small
    A toy sample, the location estimation of pendulum motion.
    We show the horizontal location as time varies, a sine curve, denoted in gray, and three estimated traces, denoted in blue, orange and cyan. They have the same $\ell_1$ mean distance to the groundtruth but have different temporal structure. Which estimated trace better describes the pendulum motion?
    The loss under different matrices is also shown in the right figure.
    Obviously, motion loss is good at answering the above question.
    }
    \label{fig:toy}
\end{figure*}

There is a critical limitation for the Minkowski Distance.
It does not consider the similarity of temporal structure between the estimated pose sequence and the groundtruth.
We illustrate this issue by a toy sample, the trace estimation of a pendulum motion.
It is similar to pose estimation, but only includes one "joint".
We compare three estimated trajectories of pendulum motion in Figure.\ref{fig:toy}.
The first trace function has a shape similar to the groudtruth.
The second one has a different tendency but still keep smoothness.
And the last curve just randomly fluctuates around the groudtruth.
Both of them have the same $\ell_1$ mean distance to the groudtruth
but have various temporal structures.
Because the Minkowski Distance is calculated independently for each moment,
it failed to examine the inner dependencies of a trajectory.

The keypoints in a pose sequence describe the human movement,
which are strongly correlated especially in the time.
Under the supervision of Minkowski Distance as loss, same as the above toy sample, it is difficult for models to learn from the motion information in the groundtruth keypoint trajectories and thus hard to obtain natural keypoints movement in the model's prediction due to the high dimensional solution space.

We address this issue by proposing \emph{motion loss}, a novel loss function that explicitly involves motion modeling into the learning.
Motion loss works by requiring the model to reconstruct the keypoint motion trajectories in addition to the task of reconstructing 3D locations of keypoints.
It evaluates the motion reconstruction quality by computing the difference between predicted keypoint locations and the ground-truth locations in the space of a specific representation called \emph{motion encoding}.
The motion encoding is built as a differentiable operator in the following manner.
We first roughly decompose a trajectory into a set of pairwise coordinate vectors with various time intervals corresponding to different time scales.
A basic differentiable binary vector operator, for instance, subtraction, inner product or cross product, is applied to each pair.
Then the obtained results are concatenated to construct the full motion encoding.
Though simple, this representation is shown in the Figure~\ref{fig:toy} (taking subtraction operator for example)
to be effective in assessing the quality of the temporal structure.
The difference in motion loss values clearly distinguishes the motion reconstruction quality of the three trajectories.
By applying it to the training of 3D pose estimation models, we also observe that motion loss can significantly improve the accuracy of 3D pose estimation.

To estimate the pose trajectories with reasonable human movements,
the 3D pose estimation model must have the capacity to model motion in both short temporal intervals and long temporal ranges,
as human actions usually have varying speeds over time.
To achieve this property we propose a novel graph convolutional network based architecture for 3D pose estimation model.
We start by repurposing an ST-GCN~\cite{yan2018spatial} model, initially proposed for skeleton-based action recognition,
to take as input 2D pose sequences and output 3D pose sequences.
Inspired by the success of U-shaped CNNs used in semantic segmentation and object detection,
we construct a similar U-shaped structure on the temporal axis of the ST-GCN~\cite{yan2018spatial} model.
The result is a new architecture, called \emph{U-shaped GCN} (UGCN),
with strong capacity in capturing both short-term and long-term temporal dependencies,
which is essential in characterizing the keypoint motion.

We experiment the motion loss and UGCN for video-based 3D pose estimation from 2D pose on two large scale 3D human pose estimation benchmarks: Human3.6M~\cite{ionescu2013human3} and MPI-INF-3DHP~\cite{mehta2017mpidata}.
We first observe a significant boost in position accuracy when the motion loss is used in training.
This corroborates the importance of motion-based supervision.
When the motion loss is combined with UGCN, our model surpasses the current state of the art models in terms of location accuracy by a large margin.
Besides improved location accuracy, we also observe that UGCN trained with the motion loss is able to produce smooth 3D sequences without imposing any smoothness constraint during training or inference.
Our model also halves the velocity error~\cite{pavllo20193d} compared with other state of the art models, which again validates the importance of having motion information in the supervision.
We provide detailed ablation study and visualization\footnote[1]{The demo video is in \href{https://www.youtube.com/watch?v=VHhsXG6OXnI&t=87s}{\textbf{\url{https://www.youtube.com/watch?v=VHhsXG6OXnI&t=87s}}}.} to further demonstrate the potential of our model.


\section{Related work}
\subsubsection{3D pose estimation.}

Before the era of deep learning,
early methods for 3D human pose estimation were based on handcraft features
~\cite{ramakrishna2012reconstructing,ionescu2013human3,ionescu2014iterated}.
In recent years, most works depend on powerful deep neural networks and achieve promising improvements,
which can be divided into two types.

In the first type, estimators
predict 3D poses from 2D images directly~\cite{li20143d,tekin2016direct,pavlakos2017coarse,tekin2016structured}.
For example,
\cite{li20143d} jointly regresses joint locations and detects body parts by sliding window on the image.
\cite{tekin2016direct} directly regresses the 3D pose from an aligned spatial-temporal feature map.
\cite{pavlakos2017coarse} predicts per voxel likelihoods for each joint based on the
stacked hourglass architecture.
\cite{tekin2016structured} utilizes an auto-encoder to learn a latent pose representation for modeling the joint dependencies.

Another typical solution build on a two-stage pipeline
~\cite{martinez2017simple,pavllo20193d,cai2019exploiting,lin2019trajectory}.
Thereon, a 2D pose sequence is firstly predicted by a 2D pose estimator from a video frame by frame
and lifted to 3D by another estimator. For instance,
\cite{martinez2017simple} proposes a simple baseline composed of several fully-connected layers,
which takes as input a single 2D pose.
\cite{pavllo20193d} generate 3D poses from 2D keypoint sequences by a temporal-convolution method.
\cite{cai2019exploiting} introduces a local-to-global network based on graph convolution.
\cite{lin2019trajectory} factorize a 3D pose sequence into trajectory bases and train a deep network to regress the trajectory coefficient matrix.

Although the appearance information is dropped in the first stage,
the data dimension is dramatically decreased as well,
which makes long-term video-based 3D pose estimation possible.
Our method also builds on the two-stage pipeline.


\subsubsection{Graph convolution.}
Modeling skeleton sequence via spatial-temporal graphs (st-graph)~\cite{yan2018spatial} and performing graph convolution thereon
have significantly boosted the performance in many human understanding tasks including
action recognition~\cite{yan2018spatial},
pose tracking~\cite{ning2019lighttrack} and
motion synthesis~\cite{yan2019convolutional}.
The designs for graph convolution mainly fall into two stream:
spectral based ~\cite{defferrard2016convolutional,kipf2016semi} and spatial based~\cite{atwood2016diffusion,niepert2016learning}.
They extended standard convolution to irregular graph domain by Fourier transformation and neighborhood partitioning respectively.
Following ~\cite{yan2018spatial}, we perform spatial graph convolution on skeleton sequences represented by st-graphs.

\section{Approach}

\begin{figure*}[t]
    \centering
    \includegraphics[width=0.98\linewidth]{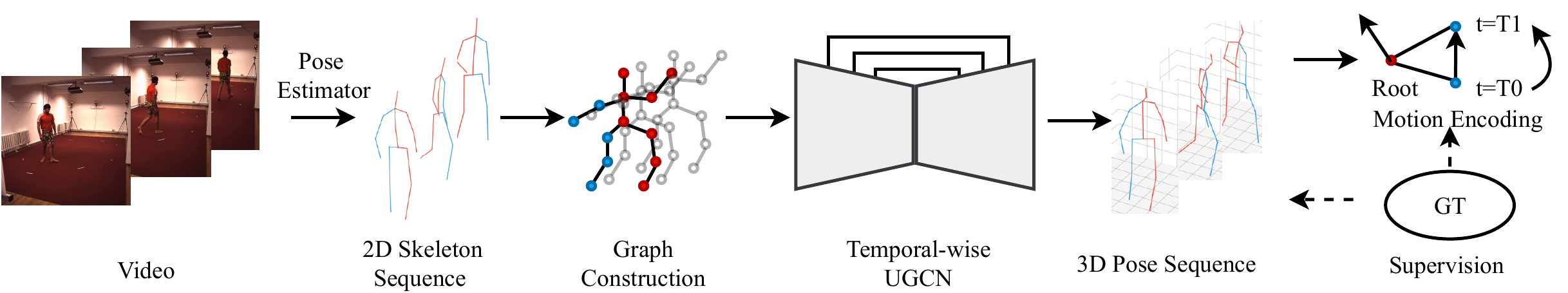}
    \caption{\small
    Overview of our proposed pipeline for
    estimating 3D poses from consecutive 2D poses.
    We structure 2D skeletons by a spatial-temporal graph and predict 3D locations via
    our U-shaped Graph Convolution Networks (UGCN).
    The model is supervised in the space of motion encoding.
    }
    \label{fig:pipeline}
\end{figure*}


Figure.~\ref{fig:pipeline} illustrates our pipeline for estimating 3D pose sequences.
Given the 2D projections of a pose sequence estimated from a video $P=\{\bm{p}_{t,j} | t=1, ..., T ; j=1, ..., M\}$,
we aim to reconstruct their 3D coordinates $S=\{\bm{s}_{t,j} | t=1, ..., T ; j=1, ..., M\}$,
where $T$ is the number of video frames, $N$ is the number of human joints,
$\bm{p}_{t,j}$ and $\bm{s}_{t,j}$ are vectors respectively representing the 2D and 3D locations of joint $j$ in the frame $t$.
We structure these 2D keypoints by a spatial-temporal graph and predict their 3D locations via
our U-shaped Graph Convolution Networks (UGCN).
The model is supervised by a multiscale motion loss and trained in an end-to-end manner.

\subsection{Motion Loss}\label{motion loss}
In this work, motion loss is defined as the distance in the space of motion.
Therefore, a motion encoder is required for projecting skeleton sequences to this space.
Though there are myriad possible designs,
we are empirically sums up a few guiding principles: differentiability, nonindependence, and multiscale.
Differentiability is the prerequisite for the end-to-end training.
And the calculation should be across time for modeling the temporal dependencies, \emph{i.e.}, nonindependence.
Since the speed of motion is different, multiscale modeling is also significant.
In this section, we introduce how we design a simple but effective encoding, named \emph{pairwise motion encoding}.


\begin{figure}[t]
    \centering
    \includegraphics[width=0.6\linewidth]{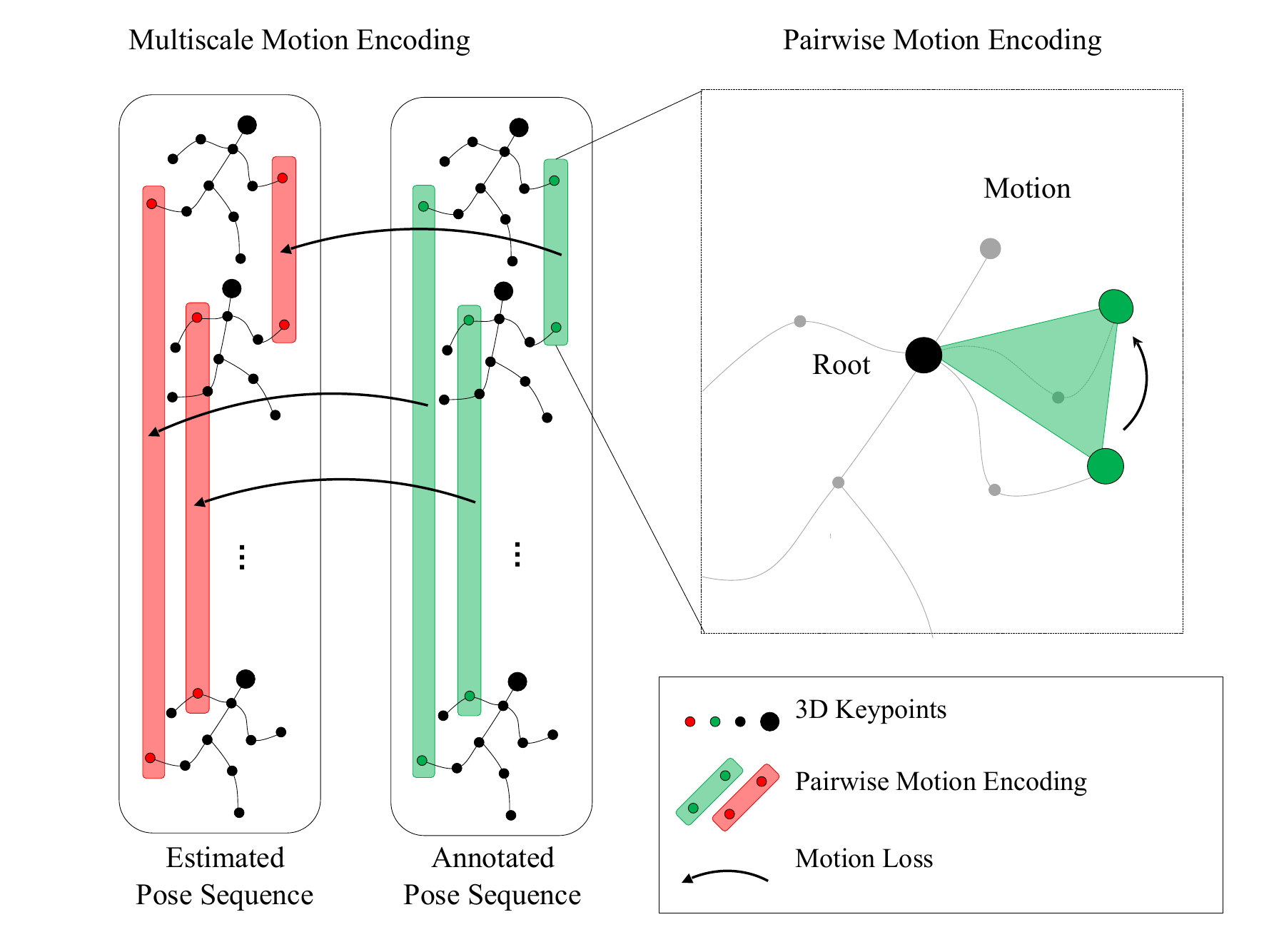}
    \caption{\small
    Motion loss. By concatenating pairwise cross product vectors between the coordinate vectors of the same joints across time with various intervals, we construct multiscale motion encoding on pose sequences.
    The motion loss requires the model to reconstruct this encoding.
    It explicitly involves motion modeling into learning.
    }
    \label{fig:teaser}
\end{figure}

\subsubsection{Pairwise motion encoding.}
We first consider the simplest case: the length of the pose sequences is $2$.
The motion encoding on the joint $j$ can be denoted as:
\begin{equation}
    \bm{m}_{j} = \bm{s}_{0,j} \star \bm{s}_{1,j},
\end{equation}
where $\star$ can be any differentiable binary vector operator,
such as subsection, inner-product and cross-prodcut.
In the common case, the pose sequence is longer.
We can expand an extra dimension in the motion encoding:
\begin{equation}
    \bm{m}_{t,j} = \bm{s}_{t,j} \star \bm{s}_{t+1,j}.
\end{equation}
Note that, this representation only models the relationship between two adjacent moments.
Since the speed of human motion has a large variation range,
it inspires us to encode human motion on multiple temporal scales:
\begin{equation}
    \bm{m}_{t,j,\tau} = \bm{s}_{t,j} \star \bm{s}_{(t+\tau),j}.
\end{equation}
where $\tau$ is the time interval. As shwon in the Figure.~\ref{fig:teaser}, to cauculate the motion loss of the full pose sequence, we compute the $ell_1$ Distance on the encoded space for all joints, moments and several time intervals. Mathematically, we have:
\begin{equation}
    L_m = \sum_{\tau\in\mathbb{T}} \sum_{t=1}^{T-\tau} \sum_{j=1}^{M} \left\| \bm{m}_{t,j,\tau} - \bm{m}_{t,j,\tau}^{gt} \right\|,
\end{equation}
where the interval set $\mathbb{T}$ includes different $\tau$ for multiple time scales.
Pairwise motion encoding decomposes a trajectory into coordinate pairs and extracts features for each pair by a differentiable operation $\star$.
As the first work to explore the supervision of motion for 3D pose estimation, intuitively, we choose the three most basic operations in the experiments: \textbf{subsection, inner-product, and cross-product}. And we conducted extensive experiments to evaluate the effectiveness of these encoding methods in Section~\ref{sec: ablation}.

\subsubsection{Loss Function.}
The motion loss only considers the second-order correlations in the formulation of pairwise motion encoding, while the absolute location information is absent.
Therefore, we add a traditional reconstruction loss term to the overall training objectives:
\begin{equation}
    L_p = \sum_{t=1}^{T} \sum_{j=1}^{M} \left\| \bm{s}_{t,j} - \bm{s}_{t,j}^{gt} \right\|_2^2.
\end{equation}
The model is supervised in an end-to-end manner with the combined loss:
 \begin{equation}
     L = L_p + \lambda L_m,
 \end{equation}
where $\lambda$ is a hyper parameter for balancing two objectives.


\subsection{U-shaped Graph Convolutional Networks}
Intuitively, the 3D pose estimator needs stronger long-term perception
for exploring the motion priors.
Besides that, keeping the spatial resolution is also required by estimating 3D pose accurately.
Therefore,
we represent the skeleton sequence as a \textit{spatial temporal graph}~\cite{yan2018spatial} to maintain their topologies,
and aggregating information by an \textit{U-shaped graph convolution network (UGCN)}.

\subsubsection{Graph Modeling}
It is an ill-posed problem to recover the 3D location of a keypoint from its 2D coordinates independently.
In general,
the information from other keypoints, especially
the neighboring ones,
play essential roles in 3D pose reconstruction.
To model the relationship with these relative keypoints,
it is natural to organize a skeleton sequence
via a \textit{spatial temporal graph} (st-graph)~\cite{yan2018spatial}.
In particular, a st-graph $G$ is determined by a node set and an edge set.
The node set $V = \{v_{t,j} | t = 1,\ldots, T, j=1,\ldots,M\}$ includes all the keypoints in a sequence of pose.
And the edge set $E$ is composed of two parts: one for connecting adjacent frames on each joint, one for the connecting endpoint of each bone in every single frame.
These edges construct the temporal dependencies and spatial configuration together.
Then, a series of graph convolution operations are conducted on this graph.

\subsubsection{Graph Convolution.}
In this work, we adopt \textit{spatial temporal graph convolution (st-gcn)}~\cite{yan2018spatial} as the basic unit
to aggregate features of nodes on a st-graph.
It can be regarded as a combination of two basic operations: a spatial graph convolution and a temporal convolution.
The temporal convolution $Conv_{t}$ is a standard convolution operation applied on the temporal dimension for each joint,
while the spatial graph convolution $Conv_{g}$ is performed on the skeleton for each time position independently.
Given an input feature map $f_{in}$, the output of two operations can be written as:
\begin{align}
    f_{s} &= Conv_{g}(f_{in}) \\
    f_{out} &= Conv_t(f_s)
\end{align}
, where $f_{s}$ is the output of the spatial graph convolution. Formally, we have:
\begin{align}
    \label{eq:graph_conv_simplify}
    f_s(v_{t,j}) &= \sum_{v_{t,i}\in B(v_{t,j})} \frac{1}{Z_{t,j}(v_{t,i})}f_{in}(v_{t,i})\cdot \mathbf{w}(l_{t,j}(v_{t,i})),
\end{align}
where $B(v_{t,j})$ is the neighbor set of node $v_{t,j}$,
$l_{t,j}$ maps a node in the neighborhood to its subset label,
$\mathbf{w}$ samples weights according to a subset label,
and $Z_{t,j}(v_{t,i})$ is a normalization term equivalent the cardinality of the corresponding subset.
Since the human torso and limbs act in very different ways,
it inspires us to give the model spatial perception for distinguishing the central joints and limbic joints.
To make spatial configuration explicit in the 3D pose estimation, we divide the neighborhood into three subsets:
\begin{equation}
l_{t,j}(v_{t,i})=
    \begin{cases}
       0 &\mbox{if $h_{t,j} = h_{t,i}$}\\
       1 &\mbox{if $h_{t,j} < h_{t,i}$}\\
       2 &\mbox{if $h_{t,j} > h_{t,i}$}
    \end{cases}
\end{equation}
, where $h_{t,i}$ is the number of hops from $v_{t,i}$ to the root node (i.e. central hip in this work).

\begin{figure*}
    \centering
    \includegraphics[width=1.0\linewidth]{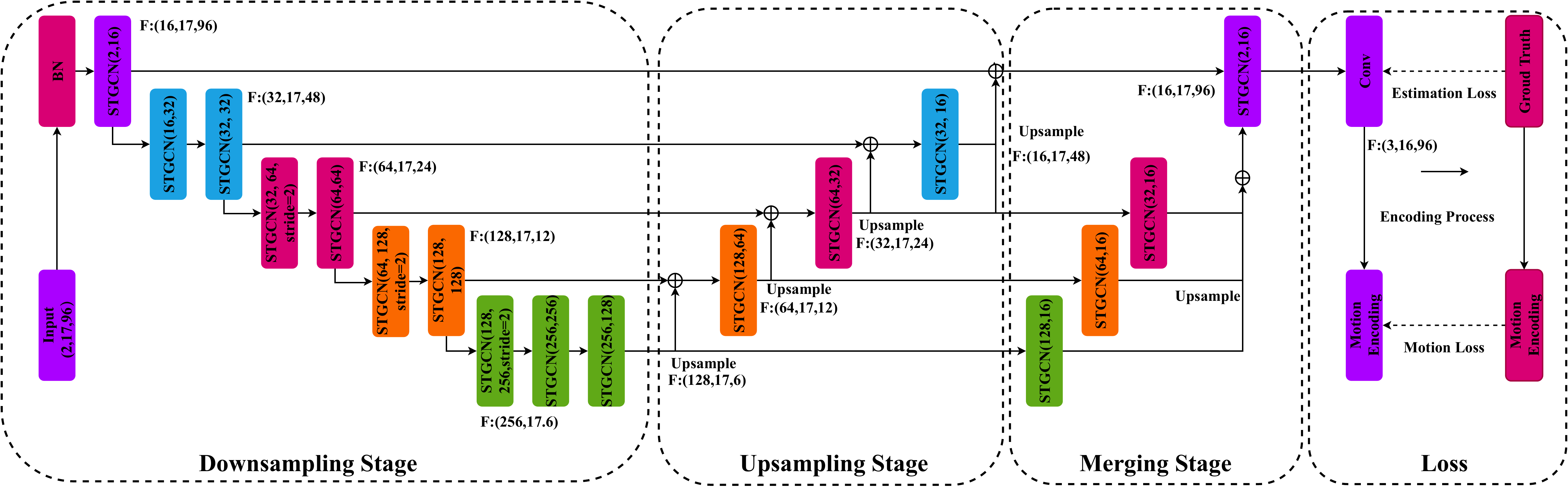}
    \caption{\small
    Network structure. We proposed a U-shaped graph convolution network (UGCN) as the backbone of our pose estimation model
    to incorporate both local and global information with a high resolution.
    This network consists of three stages:
    downsampling, upsampling and merging.
    The network first aggregates long-range information by temporal pooling operations in the downsampling stage.
    And then recovers the resolution by upsampling layers.
    To keep the low-level information,
    the features in the downsampling stage are also added to the upsample branch by some shortcuts.
    Finally, the multi-scale feature maps are merged to predicted 3D skeletal joints.
    In this way, UGCN incorporates both short-term and long-term information, making it an ideal fit for the supervision of the motion loss.
    }
    \label{fig:network}
\end{figure*}

\subsubsection{Network structure.}

As shown in Figure~\ref{fig:network},
the basic units for building networks are st-gcn blocks,
which include five basic operations:
a spatial graph convolution,
a temporal convolution,
a batch normalization,
a dropout and an activation function ReLu.
Our networks are composed of three stages: downsampling, upsampling, and merging.

In the downsampling stage,
we utilize 9 st-gcn blocks for aggregating temporal features.
In addition, we set $stride=2$ for the second, fourth, sixth, and eighth st-gcn blocks to increase the receptive field in the time dimension.
This stage embeds the global information of the full skeleton sequence.

The upsampling stage contains four st-gcn blocks.
Each block is followed by an upsampling layer.
Thanks to the regular temporal structure in st-graph, the upsampling in the time dimension can be simply implemented with the following
formula:
\begin{align}
    f_{up}(v_{t,j}) &= f_{in}(v_{t',i}),
\end{align}
where $t' = \left \lfloor \frac{t}{2} \right \rfloor$.
With successive upsampling operations,
the temporal resolution gradually recovers
and the global information spread to the full graph.
Since the 2D inputs are projections of 3D outputs,
the low-level information may provide strong geometric constraints for estimating 3D pose.
It motivated us to keep low-level information in the networks.
Thus, we add features in the first stage to the upsampling stage with the same temporal resolution.

In the merging stage, the feature maps with various time scales in the second stage are transformed to the same shape and fused to obtain the final embedding.
Obviously, this embedding contains abundant information on multiple temporal scales.

In the end,
the 3D coordinate for each keypoint is estimated by a st-gcn regressor.
This model is supervised by the motion loss in an end-to-end manner.
Other details have been depicted in the Figure~\ref{fig:network}.

\subsubsection{Training \& inference.}
We use st-gcn blocks with the temporal kernel size of $5$ and the dropout rate of $0.5$ as our basic cells to construct a UGCN.
The networks take as input a 2D pose sequence with $96$ frames.
We perform horizontal flip augmentation at the time of training and testing.
It is supervised by a motion loss with $\lambda=1$.
We optimize the model using Adam for $110$ epochs with the batch size of $256$ and the initial learning rate of $10^{-2}$. We decay the learning rate by $0.1$ after $80$, $90$ and $100$ epochs.
To avoid the overfitting, we set the weight decay factor to $10^{-5}$ for parameters of convolution layers.

In the inference stage,
we apply the sliding window algorithm with the step length of $5$ to estimate a variable-length
pose sequence with fixed input length,
and average all results on different time positions.
\section{Experiments}

We evaluate models on two large-scale datasets for 3D pose estimation: Human3.6M and MPI-INF-3DHP.
In particular, we first perform detailed ablation study on the Human3.6M dataset
to examine the effectiveness of the proposed components.
To exclude the interference of 2D pose estimator, all experiments in this ablation study take 2D ground truth as input.
Then, we compare the estimated results of UGCN with other state-of-the-art methods on two datasets.
All experiments are conducted on PyTorch tools with one single TITANX GPU.

\subsection{Dataset}
\textbf{Human3.6M:}
Human3.6M~\cite{h36m_pami} is a large-scale indoor dataset for 3D human pose estimation.
This widely used dataset consists of 3.6 million images which are captured from 4 different cameras.
There are 11 different subjects and 15 different actions in this dataset, such as ``Sitting'', ``Walking'', and``Phoning''.
The 3D ground truth and all parameters of the calibrated camera systems are provided in this dataset. %
However, we do not exploit the camera parameters in the proposed approach.
Following the recent works, we utilize (S1, S5, S6, S7, S8) for training and (S9, S11) for testing.
The video from all views and all actions are trained by a single model.
For this dataset, we conduct ablation studies based on the ground truth of 2D skeleton.
Besides that, we also report the results of our approach taking as input predicted 2D poses.
from widely used pose estimators.

\begin{table}
    \scriptsize
    \centering
    \setlength{\tabcolsep}{0.6em}
    \caption {Performance of our UGCN model supervised by motion loss with different basic operators and time intervals. The empty set $\varnothing$ denotes that the motion loss is not utilized.
    The best MPJPE is achieved by the cross product operator with interval of $12$.}
    \label{tab: interval}
    \begin{tabular}{@{}l|ccccccccc@{}}
    \toprule
    Interval set $\mathbb{T}$  & $\varnothing$ & \{2\} & \{4\} & \{8\} & \{12\} & \{16\} & \{24\} & \{36\} & \{48\}   \\
    \midrule
    Subtraction & 32.0 &31.4 & 30.8 & 29.7 & \textbf{28.9} & 29.3 & 30.6 & 31.8 & 32.8   \\
    Inner Product & 32.0&31.8 & 31.7 & 31.0 & 30.2 & \textbf{29.8} & 31.2 & 32.6 & 33.7   \\
    Cross Product & 32.0&31.2 & 30.4 & 28.2 & \textbf{27.1} & 28.3 & 30.2 & 31.6 & 32.7   \\
    \bottomrule
    \end{tabular}
\end{table}
\begin{table}
\scriptsize
\setlength{\tabcolsep}{0.6em}
\centering
\caption {We select the $4$ best time intervals according to the Table.\ref{tab: interval},
and add them to the interval set one by one. More keypoint pairs with different intervals involve the calculation of mothion encoding.
The MPJPE is improved in this process.}
\label{tab: interval2}
\begin{tabular}{@{}c|cccc|c|c @{}}
\toprule
Operator  & $\tau=8$ &  $\tau=12$ &  $\tau=16$ &  $\tau=24$& \# Time Scales& MPJPE(mm)   \\
\midrule
  Cross Product &              & $\checkmark$ &  &  & 1 &27.1 \\
  Cross Product & $\checkmark$ & $\checkmark$ &  &  & 2 &26.3 \\
  Cross Product & $\checkmark$ & $\checkmark$ & $\checkmark$  &  & 3 &25.7  \\
  Cross Product & $\checkmark$ & $\checkmark$ & $\checkmark$  & $\checkmark$  & 4 &\textbf{25.6} \\
\bottomrule
\end{tabular}
\end{table}

\noindent
\textbf{MPI-INF-3DHP:}
MPI-INF-3DHP~\cite{mono-3dhp2017} is a recently released 3D human pose estimation dataset.
And this dataset is captured in both indoor environment and in-the-wild outdoor environment.
Similar to Human3.6M, this dataset also provides videos from different cameras, subjects, and actions.

\subsection{Evaluation Metric}
For both Human3.6M and MPI-INF-3DHP dataset, we report the \textit{mean per joint position error(MPJPE)}~\cite{lin2019trajectory,pavllo20193d,cai2019exploiting} as the evaluation metric.
In general, there are two protocals, \textit{Protocal-1} and \textit{Protocal-2}, used in the previous works to evaluate 3D pose estimation.
Metric Protocol-1 first aligns the root joint(central hip) and then calculates the average Euclidean distance of the estimated joints.
While in the Protocol-2, the estimated results are further aligned to the ground truth via a rigid transformation before computing distance.

In MPI-INF-3DHP, we evaluate models under two additional metrics.
The first one is the area under the curve (AUC)~\cite{zimmermann2017learning} on the percentage of correct keypoints(PCK) score for different error
thresholds. Besides, PCK with the threshold of $150mm$ is also reported.

\subsection{Ablation Study}~\label{sec: ablation}
In this section, we demonstrate the effectiveness of the proposed UGCN and our motion loss on the Human3.6M dataset.
Experiments in this section directly take 2D ground-truth as input to eliminate the interference of 2D pose estimator.

\subsubsection{Effect of motion loss.}

    We start our ablation study from observing the impact of the temporal interval $\tau$ in the single scale motion loss.
    In other words, the interval set for motion loss has only one element.
    The value of this element controls the temporal scale of motion loss.
    We conduct experiments on three binary operators proposed in Section~\ref{motion loss}, \emph{i.e.} subtraction, inner production and cross production.

    As shown in Table~\ref{tab: interval},
    the cross production achieves the lowest MPJPE error with almost all temporal intervals.
    Besides, the MPJPE error decrease first and then increase,
    and reduce the error by $4.9mm$ (from 32.0 to 27.1) with the time interval of $12$ and the cross production encoding.
    There are two observations.
    First, compared to the result without motion term (denoted as $\varnothing$), even the temporal interval is large (24 frames),
    the performance gain is still positives. It implies that motion prior is not momentary.
    And the model might need long-term perception for better capturing the motion information.
    Second, motion loss boosts the performance with temporal interval $\tau$ in a large variation range (2$\sim$36 frames),
    which means the time scale of motion priors is also various.

    Thus, it is reasonable to adopt motion loss with multiple time intervals.
    We select four best $\tau$ as candidates and adopt the most effective binary operator in Table.~\ref{tab: interval}, cross production.
    The experimental results have been depicted in Table.~\ref{tab: interval2}.
    Under the supervision of multiscale motion loss,
    our model decrease the MPJPE by $1.5mm$ (27.1 $\rightarrow$ 25.6).

\begin{table}[htb]
\scriptsize
\setlength{\tabcolsep}{0.39em}
\centering
\caption {We remove all downsampling and upsampling operations from the standard UGCN, and add them back pair by pair. The MPJPE performance of our system increases remarkably in this process. With motion loss, the achieved gain is even large.}
\label{tab: downsample}
\begin{tabular}{@{}l|cccc|c@{}}
\toprule
\# Downsample \& Upsample   & 0& 1 & 2 & 4 &$\Delta$\\
\midrule
UGCN w/o Motion Loss      & 38.6 & 37.2 & 36.9 & 32.0 &6.6   \\
UGCN + Motion Loss ($\mathbb{T}=\{12\}$) & 36.9 & 34.8 & 33.7 & \textbf{27.1} & \textbf{9.8}\\
\midrule
$\Delta$ & 1.7 & 2.4 & 3.2 & 4.9& -\\
\bottomrule
\end{tabular}
\end{table}

\begin{table}[htb]
    \scriptsize
    \setlength{\tabcolsep}{0.39em}
    \centering
    \caption {We explore the importance of each individual component by removing them from standard setting.
        The increased MPJPE error for each module is listed below.}
    \label{tab: graph and merge}
    \begin{tabular}{@{}lcc@{}}
    \toprule
    Backbone    &  MPJPE(mm)& $\Delta$ \\
    \midrule
    UGCN & 32.0 & - \\
    UGCN w/o Spatial Graph  & 39.2 &  7.2\\
    UGCN w/o Merging Stage  &32.5 & 0.5 \\
    \midrule
    UGCN                  + Motion Loss  & 25.6 & - \\
    UGCN + Motion Loss w/o Merging Stage   & 28.4 & 2.8 \\
    \bottomrule
    \end{tabular}
\end{table}

\subsubsection{Design choices in UGCN.}
We first examine the impact of the U-shaped architecture.
We remove all downsampling and upsampling operations from the standard UGCN,
and add them back pair by pair.
The experimental results have been depicted in Table.~\ref{tab: downsample}.
It can be seen that U-shaped structure brings significant improvement ($6.6mm$) to UGCN.
This structure even leads to a larger performance gain ($9.8mm$) with the supervision of motion loss.
And the gap caused by motion loss is growing with the increasing number of downsampling and upsampling.
These results validate our assumption: the motion loss requires long-term perception.

We also explore other design choices in the UGCN. As shown in Table.~\ref{tab: graph and merge},
the spatial configuration bring $7.2mm$ improvement.
Removing the merging stage only slightly enlarge the error.
However, when the model is supervised by motion loss, the performance drop is more remarkable ($0.5mm$ vs. $2.8mm$).
That is to say, multiscale temporal information is important to the learning of motion prior.

    \begin{table}[htb]
        \scriptsize
        \setlength{\tabcolsep}{0.39em}
    \centering
    \caption { The MPJPE performance of our system with different supervision.
    Combining motion loss functions with different basic operators does not bring obvious improvement.}
    \label{tab: compare2}
    \begin{tabular}{@{}lccc @{}}
    \toprule
    Loss Function & Interval set $\mathbb{T}$~~ & MPJPE(mm) & $\Delta$ \\
    \midrule
    -       & $\varnothing$& 32.0 & - \\
    Derivative loss~\cite{rayat2018exploiting}   &$\{1\}$& 31.6 & 0.4 \\
    \midrule
    Cross product   &$\{12\}$& 27.1 & 4.9 \\
    Subtraction+ Cross product   &$\{12\}$& 27.1 & 4.9 \\
    Subtraction + Inner + Cross product & $\{12\}$ & 27.1 & 4.9 \\
    %
    \bottomrule
    \end{tabular}
\end{table}

\subsubsection{Design choices in motion loss.}

The formula of offset encoding is similar to the Derivative Loss~\cite{rayat2018exploiting}
which regularizes the joint offset between adjacent frames.
This loss is under the the hypothesis that the motion is smooth between the neighborhood frames.
We extend it to our motion loss formulation.
Since only short-term relation is considered, the improvement achieved by Derivative Loss is minor.
Then we compare the results of our method supervised by the motion loss with different combination of the proposed binary operators.
The results have been shown in Table.~\ref{tab: compare2}.
The combination of these three representations is not able to bring any improvement.
Therefore, we adopt cross production as the pairwise motion encoder in the following experiments.
\begin{table}[t]
    \tiny
    \setlength{\tabcolsep}{0.15em}
    \centering
    \caption {
        Results showing the errors action-wise on Human3.6M under Protocol-1 and Protocol-2.
        (CPN) and (HRNET) respectively indicates the model trained on 2D poses estimated by CPN~\cite{chen2018cascaded},
        and HR-Net~\cite{Sun_2019_CVPR}. $\dagger$ means the methods adopt the same refine module as~\cite{cai2019exploiting}.
        }
    \label{tab: h35m sota}
    \begin{tabular}[width=0.9\linewidth]{@{}l| ccccccccccccccc |c   @{}}
    \toprule
    \textbf{\emph{Protocol 1}} & Dir. & Disc. & Eat. & Greet   & Phone & Photo & Pose & Purch. & Sit   & SitD. & Somke & Wait & WalkD.  & Walk & WalkT. & Ave. \\
    \midrule
    Mehta~\cite{mehta2017monocular} &57.5 & 68.6 &59.6 &67.3 &78.1& 82.4 &56.9& 69.1& 100.0& 117.5& 69.4& 68.0& 55.2 &76.5& 61.4& 72.9 \\
    Pavlakos~\cite{pavlakos2017coarse} & 67.4& 71.9& 66.7& 69.1& 72.0& 77.0& 65.0& 68.3& 83.7& 96.5& 71.7& 65.8& 74.9& 59.1& 63.2& 71.9 \\
    Zhou~\cite{zhou2017towards} & 54.8& 60.7& 58.2& 71.4& 62.0& 65.5& 53.8& 55.6& 75.2& 111.6& 64.1& 66.0& 51.4& 63.2& 55.3& 64.9 \\
    Martinez~\cite{martinez2017simple}& 51.8& 56.2& 58.1& 59.0& 69.5& 78.4& 55.2& 58.1& 74.0& 94.6& 62.3& 59.1& 65.1& 49.5& 52.4& 62.9\\
    Sun~\cite{sun2017compositional} & 52.8& 54.8& 54.2& 54.3& 61.8& 67.2& 53.1& 53.6& 71.7& 86.7& 61.5& 53.4& 61.6& 47.1& 53.4& 59.1 \\
    Fang~\cite{fang2018learning} & 50.1& 54.3& 57.0& 57.1& 66.6& 73.3& 53.4& 55.7& 72.8& 88.6& 60.3& 57.7& 62.7& 47.5& 50.6& 60.4 \\
    Pavlakos~\cite{pavlakos2018ordinal} & 48.5& 54.4& 54.4& 52.0& 59.4& 65.3& 49.9& 52.9& 65.8& 71.1& 56.6& 52.9& 60.9& 44.7& 47.8& 56.2 \\
    Lee~\cite{lee2018propagating} & 43.8& 51.7& 48.8& 53.1& 52.2& 74.9& 52.7& 44.6& 56.9& 74.3& 56.7& 66.4& 68.4& 47.5& 45.6& 55.8 \\
    \midrule
    Zhou~\cite{zhou2016sparseness} & 87.4& 109.3& 87.1& 103.2& 116.2& 143.3& 106.9& 99.8& 124.5& 199.2& 107.4& 118.1& 114.2& 79.4& 97.7& 113.0 \\
    Lin~\cite{lin2017recurrent} & 58.0& 68.2& 63.3& 65.8& 75.3& 93.1& 61.2& 65.7& 98.7& 127.7& 70.4& 68.2& 72.9& 50.6& 57.7& 73.1 \\
    Hossain~\cite{rayat2018exploiting} & 48.4& 50.7& 57.2& 55.2& 63.1& 72.6& 53.0& 51.7& 66.1& 80.9& 59.0& 57.3& 62.4& 46.6& 49.6& 58.3 \\
    Lee~\cite{lee2018propagating}(F=3) & 40.2& 49.2& 47.8& 52.6& 50.1& 75.0& 50.2& 43.0& 55.8& 73.9& 54.1& 55.6& 58.2& 43.3& 43.3& 52.8 \\
    Dabral~\cite{dabral2018learning} & 44.8& 50.4& 44.7& 49.0& 52.9& 61.4& 43.5& 45.5& 63.1& 87.3& 51.7& 48.5& 52.2& 37.6& 41.9& 52.1 \\
    Pavllo~\cite{pavllo20193d} & 45.2& 46.7& 43.3& 45.6& 48.1& 55.1& 44.6& 44.3& 57.3& 65.8& 47.1& 44.0& 49.0& 32.8& 33.9& 46.8\\
    Cai~\cite{cai2019exploiting}$\dagger$ & 44.6& 47.4& 45.6& 48.8& 50.8& 59.0& 47.2& 43.9& 57.9& 61.9& 49.7& 46.6& 51.3& 37.1& 39.4& 48.8 \\
    Lin~\cite{lin2019trajectory} & 42.5& 44.8& 42.6& 44.2& 48.5& 57.1& 42.6& 41.4& 56.5& 64.5& 47.4& 43.0& 48.1& 33.0& 35.1& 46.6 \\
    \midrule
    UGCN(CPN)& 41.3 & 43.9 & 44.0 & 42.2 & 48.0 & 57.1 & 42.2 & 43.2 & 57.3 & 61.3 & 47.0 & 43.5 & 47.0 & 32.6 & 31.8 & 45.6\\
    UGCN(CPN)$\dagger$ & 40.2 & 42.5 & 42.6 & 41.1 & 46.7 & 56.7 & 41.4 & 42.3 & 56.2 & 60.4 & 46.3 & 42.2 & 46.2 & 31.7 & 31.0 & 44.5 \\
   	UGCN(HR-Net) & \textbf{38.2} & \textbf{41.0} & \textbf{45.9} & \textbf{39.7} & \textbf{41.4} &\textbf{51.4} & \textbf{41.6} & \textbf{41.4} & \textbf{52.0} & \textbf{57.4} & \textbf{41.8} & \textbf{44.4} & \textbf{41.6} & \textbf{33.1} & \textbf{30.0} & \textbf{42.6} \\
    \bottomrule
    \toprule
    \textbf{\emph{Protocol 2}} & Dir. & Disc. & Eat. & Greet   & Phone & Photo & Pose & Purch. & Sit   & SitD. & Somke & Wait & WalkD.  & Walk & WalkT. & Ave. \\
    \midrule
    Martinez~\cite{martinez2017simple} & 39.5& 43.2& 46.4& 47.0& 51.0& 56.0& 41.4& 40.6& 56.5& 69.4& 49.2& 45.0& 49.5& 38.0& 43.1& 47.7 \\
    Sun~\cite{sun2017compositional} & 42.1& 44.3& 45.0& 45.4& 51.5& 53.0& 43.2& 41.3& 59.3& 73.3& 51.0& 44.0& 48.0& 38.3& 44.8& 48.3 \\
    Fang~\cite{fang2018learning} & 38.2& 41.7& 43.7& 44.9& 48.5& 55.3& 40.2& 38.2& 54.5& 64.4& 47.2& 44.3& 47.3& 36.7& 41.7& 45.7 \\
    Lee~\cite{lee2018propagating} & 38.0& 39.3& 46.3& 44.4& 49.0& 55.1& 40.2& 41.1& 53.2& 68.9& 51.0& 39.1& 56.4& 33.9& 38.5& 46.2 \\
    Pavlakos~\cite{pavlakos2018ordinal} & 34.7& 39.8& 41.8& 38.6& 42.5& 47.5& 38.0& 36.6& 50.7& 56.8& 42.6& 39.6& 43.9& 32.1& 36.5& 41.8 \\
    \midrule
    Hossain~\cite{rayat2018exploiting} & 35.7& 39.3& 44.6& 43.0& 47.2& 54.0& 38.3& 37.5& 51.6& 61.3& 46.5& 41.4& 47.3& 34.2& 39.4& 44.1 \\
    Pavllo~\cite{pavllo20193d} & 34.1& 36.1& 34.4& 37.2& 36.4& 42.2& 34.4& 33.6& 45.0& 52.5& 37.4& 33.8& 37.8& 25.6& 27.3& 36.5 \\
    Dabral~\cite{dabral2018learning} & \textbf{28.0}& \textbf{30.7}& 39.1& 34.4& 37.1& 44.8& \textbf{28.9}& \textbf{31.2}& \textbf{39.3}& 60.6& 39.3& \textbf{31.1}& 37.8& 25.3& 28.4& 36.3 \\
    Cai~\cite{cai2019exploiting}$\dagger$ & 35.7& 37.8& 36.9& 40.7& 39.6& 45.2& 37.4& 34.5& 46.9& 50.1& 40.5& 36.1& 41.0& 29.6& 33.2& 39.0 \\
    Lin~\cite{lin2019trajectory} & 32.5& 35.3& 34.3& 36.2& 37.8& 43.0& 33.0& 32.2& 45.7& 51.8& 38.4& 32.8& 37.5& 25.8& 28.9& 36.8 \\
    \midrule
    UGCN(CPN)& 32.9 & 35.2 & 35.6 & 34.4 & 36.4 & 42.7 & 31.2 & 32.5 & 45.6 & 50.2 & 37.3 & 32.8 & 36.3 & 26.0 & 23.9 & 35.5 \\
    UGCN(CPN)$\dagger$ & 31.8 & 34.3 & 35.4 & 33.5 & 35.4 & 41.7 & 31.1 & 31.6 & 44.4 & 49.0 & 36.4 & 32.2 & 35.0 & \textbf{24.9} & 23.0 & 34.5 \\
   UGCN(HR-Net)& 28.4& 32.5 & \textbf{34.4} & \textbf{32.3} & \textbf{32.5} & \textbf{40.9} & 30.4 & 29.3 & 42.6 & \textbf{45.2} & \textbf{33.0} & 32.0& \textbf{33.2} & \textbf{24.2} & \textbf{22.9} & \textbf{32.7} \\
    \bottomrule
    \end{tabular}
\end{table}

\begin{table}[htb]
    \tiny
    \setlength{\tabcolsep}{0.25em}
    \centering
    \caption {To exclude the interference of 2D pose estimator, we compare our models and state-of-the-arts trained on ground truth 2D pose. Results showing the action-wise errors on Human3.6M under Protocol-1.}
    \label{tab: h35m sota gt}
    \begin{tabular}[width=0.9\linewidth]{@{}l| ccccccccccccccc |c   @{}}
    \toprule
    \textbf{\emph{Protocol 1 (GT)}} & Dir. & Disc. & Eat. & Greet   & Phone & Photo & Pose & Purch. & Sit   & SitD. & Somke & Wait & WalkD.  & Walk & WalkT. & Ave. \\
    \midrule
    Pavlakos~\cite{pavlakos2018ordinal} & 47.5& 50.5& 48.3& 49.3& 50.7& 55.2& 46.1& 48.0& 61.1& 78.1& 51.05& 48.3& 52.9& 41.5& 46.4& 51.9 \\
    Martinez~\cite{martinez2017simple} & 37.7& 44.4& 40.3& 42.1& 48.2& 54.9& 44.4& 42.1& 54.6& 58.0& 45.1& 46.4& 47.6& 36.4& 40.4& 45.5\\
    Hossain~\cite{rayat2018exploiting} & 35.7& 39.3& 44.6& 43& 47.2& 54.0& 38.3& 37.5& 51.6& 61.3& 46.5& 41.4& 47.3& 34.2& 39.4& 44.1 \\
    Lee~\cite{lee2018propagating} & 34.6& 39.7& 37.2& 40.9& 45.6& 50.5& 42.0& 39.4& 47.3& 48.1& 39.5& 38.0& 31.9& 41.5& 37.2& 40.9 \\
    Pavllo~\cite{pavllo20193d} & - &- & -& -& -& -& -& -& -& -& -&-& -& -&- & 37.2 \\
    Cai~\cite{cai2019exploiting} & 32.9& 38.7& 32.9& 37.0& 37.3& 44.8& 38.7& 36.1& 41.0& 45.6& 36.8& 37.7& 37.7& 29.5& 31.6& 37.2\\
    Lin~\cite{lin2019trajectory} & 30.1 & 33.7 & 28.7 & 31.0 & 33.7 & 40.1 & 33.8 & 28.5 & 38.6 & 40.8 & 32.4 & 31.7 & 33.8 & 25.3 & 24.3 & 32.8 \\
    \midrule
    UGCN & \textbf{23.0} & \textbf{25.7} & \textbf{22.8} & \textbf{22.6} & \textbf{24.1} & \textbf{30.6} & \textbf{24.9} & \textbf{24.5} & \textbf{31.1} & \textbf{35.0} & \textbf{25.6} & \textbf{24.3} & \textbf{25.1} & \textbf{19.8} & \textbf{18.4} & \textbf{25.6} \\
    \bottomrule
    \end{tabular}
\end{table}
\begin{table}
    \tiny
    \setlength{\tabcolsep}{0.2em}
    \centering
    \caption {Results show the velocity error of our methods and other state-of-the-arts on Human3.6M.
        Our result without motion loss is denoted as (*).}
    \label{tab: h35m sota 3}
    \begin{tabular}[width=0.9\linewidth]{@{}l| ccccccccccccccc |c   @{}}
    \toprule
    MPJVE & Dir. & Disc. & Eat. & Greet   & Phone & Photo & Pose & Purch. & Sit   & SitD. & Somke & Wait & WalkD.  & Walk & WalkT. & Ave. \\
    \midrule
    Pavllo~\cite{pavllo20193d} & 3.0& 3.1& 2.2& 3.4& 2.3& 2.7& 2.7& 3.1& 2.1& 2.9& 2.3& 2.4& 3.7& 3.1& 2.8& 2.8 \\
    Lin~\cite{lin2019trajectory} & 2.7 & 2.8 & 2.1 & 3.1 & 2.0 & 2.5 & 2.5 & 2.9 & 1.8 & 2.6 & 2.1 & 2.3 & 3.7 & 2.7 & 3.1 & 2.7 \\
    \midrule
    UGCN(CPN)*& 3.5 & 3.6 & 3.0 & 3.9 & 3.0 & 3.4 & 3.2 & 3.6 & 2.9 & 3.7  & 3.0 & 3.1 & 4.2 & 3.4 & 3.7 & 3.4 \\
    UGCN(CPN)& 2.3 & 2.5 & 2.0 & 2.7 & 2.0 & 2.3 & 2.2 & 2.5 & 1.8 & 2.7 & 1.9 & 2.0 & 3.1 & 2.2 & 2.5 & 2.3 \\
    UGCN(GT) & 1.2 & 1.3 & 1.1 & 1.4 & 1.1 & 1.4 & 1.2 & 1.4 & 1.0 & 1.3 & 1.0 & 1.1 & 1.7 & 1.3 & 1.4 & 1.4\\
    \bottomrule
    \end{tabular}
\end{table}
\begin{table}
    \scriptsize
    \setlength{\tabcolsep}{1.3em}
    \centering
    \caption {Comparison with previous work on the MPI-INF-3DHP dataset. The bold-faced
    numbers represent the best, while underlined numbers represent
    the second best.}
    \label{tab: mpii3}
    \begin{tabular}{@{}l|ccc@{}}
    \toprule
    Method & PCK[$\uparrow$] & AUC[$\uparrow$] & MPJPE(mm)[$\downarrow$] \\
    \midrule
    Mehta~\cite{mehta2017monocular} & 75.7 & 39.3 & - \\
    Mehta(ResNet=50)~\cite{mehta2017vnect} & 77.8 & 41.0 & - \\
    Mehta(ResNet=101)~\cite{mehta2017vnect} & 79.4 & 41.6 & - \\
    Lin(F=25)~\cite{lin2019trajectory} & 83.6 & 51.4 & 79.8 \\
    Lin(F=50)~\cite{lin2019trajectory} & 82.4 & 49.6 & 81.9 \\
    \midrule
    UGCN w/o Motion Loss & \underline{84.2} & \underline{54.2} & \underline{76.7}\\
    UGCN & \textbf{86.9} & \textbf{62.1} & \textbf{68.1} \\
    \bottomrule
    \end{tabular}
\end{table}

\begin{figure*}[htb]
    \centering
    \includegraphics[width=0.9\linewidth]{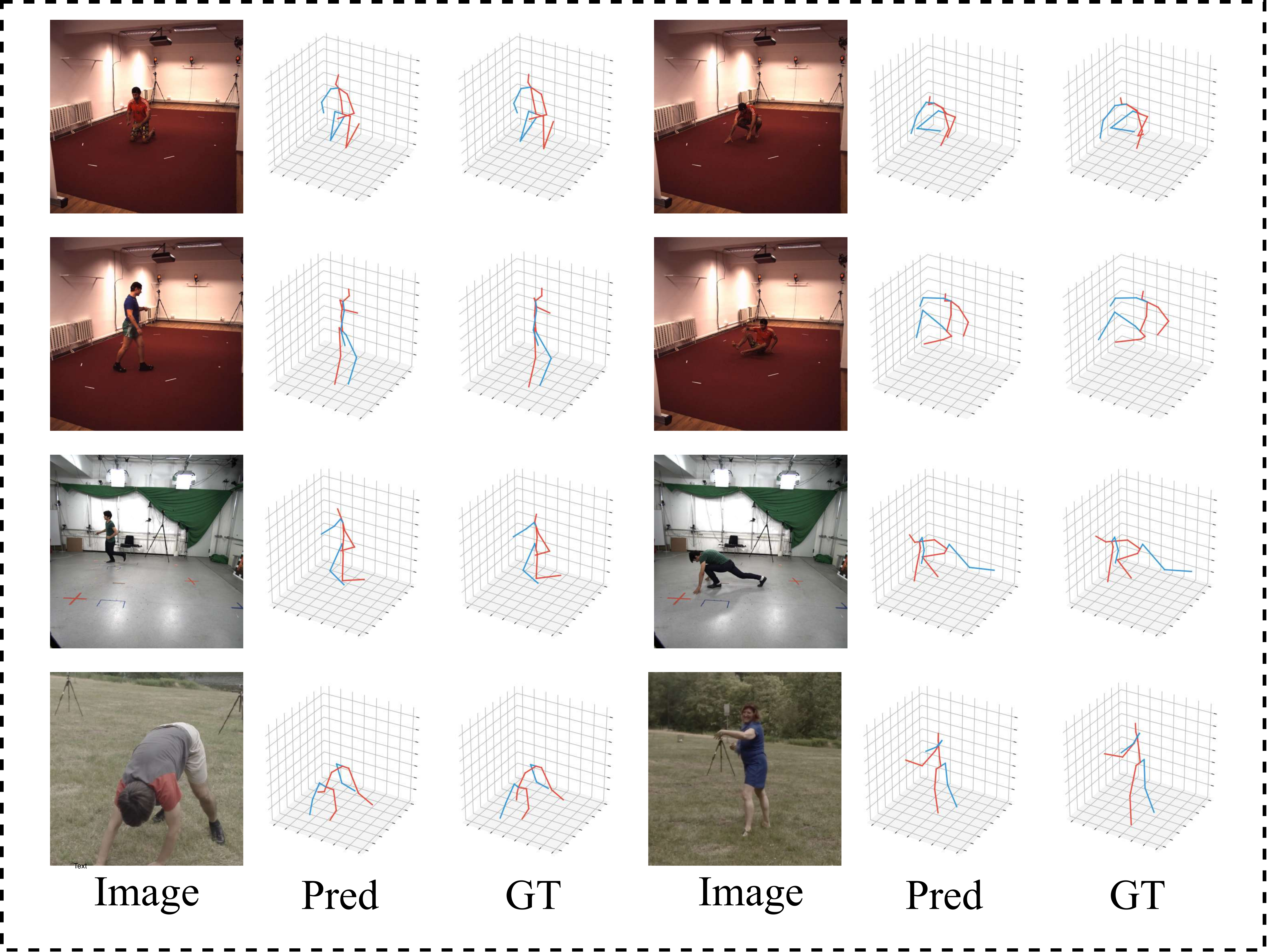}
    \caption{\small
    Visulation results of our full system on Human3.6M and MPI-INF-3DHP.
    }
    \label{fig:visulation}
\end{figure*}

\subsection{Comparison with state-of-the-art}

\subsubsection{Results on Human3.6M}
In this section, we compare the proposed approach to several \emph{state-of-the-art}
algorithms in monocular 3D pose estimation from an agnostic camera on Human3.6M dataset.
%
%
%
We trained our model on 2D poses predicted by cascaded pyramid network (CPN)~\cite{chen2018cascaded}.
It is the most typical 2D estimator used in previous works.
The results on two protocols are shown in the Table~\ref{tab: h35m sota}. 
%
As shown in the table,
our method achieves promising results on Human3.6
under two metrics(45.6 MPJPE on \textit{Protocal 1}
and 35.5 P-MPJPE on \textit{Protocal 2}) which surpass all other baselines.
We also examine the result on a more powerful 2D pose estimator HR-Net~\cite{Sun_2019_CVPR}.
It further brings roughly $3mm$ MPJPE improvement.

Several state-of-the-arts report their results on 2D ground-truth to explore their upper bound in 3D pose estimation.
The results are illustrated in the Table~\ref{tab: h35m sota gt}.
It can be seen that our method achieves the best performance (25.6 MPJPE) outperforming all other methods with the ground-truth input.



Following~\cite{pavllo20193d}, we evaluate the dynamic quality of predicted 3D pose sequences by Mean per Joint Velocity Error(MPJVE).
This metric measures the smoothness of predicted pose sequences.
As shown in Table~\ref{tab: h35m sota 3},
with motion loss, our method significantly reduces the MPJVE by 32\% (from $3.4mm$ to $2.3mm$) and outperforms other baselines.



\subsubsection{Results on MPI-INF-3DHP}
%
We compare the results of PCK, AUC, and MPJPE against the other state-of-the-art methods on MPI-INF-3DHP dataset with the input of groud truth 2d skeleton sequences.
As shown in Table~\ref{tab: mpii3}, our approach achieves a significant improvement against other methods.
Our method finally achieves 86.9 PCK, 62.1 AUC and 68.1 MPJPE on this dataset.
The proposed motion loss significantly improves the accuracy and reduces the error.

\subsection{Visualization results}
The qualitative results on Human3.6M and MPI-INF-3DHP are shown in Figure~\ref{fig:visulation}.
We choose samples with huge movements and hard actions to show the effectiveness of our system.
More visualization results comparing with other previous works can be find in the appendix section.

\section{Conclusion}
In this work, we propose a novel objective function, motion loss.
It explicitly involves motion modeling into learning.
To better optimize model under the supervision of motion loss, the 3D pose estimation should have a long-term perception of pose sequences.
It motivated us to design an U-shaped model to capture both short-term and long-term temporal dependencies.
On two large datasets, the proposed UGCN with motion loss
achieves state-of-the-art performance.
The motion loss may inspire other skeleton-based tasks such as action forecasting, action generation and pose tracking.


%
%
\bibliographystyle{splncs04}
\bibliography{main.bib}

\clearpage

\section*{Appendix A}
\subsection*{Impact of 2D Pose Estimators}

\begin{figure*}[htb]
    \centering
    \includegraphics[width=0.98\linewidth]{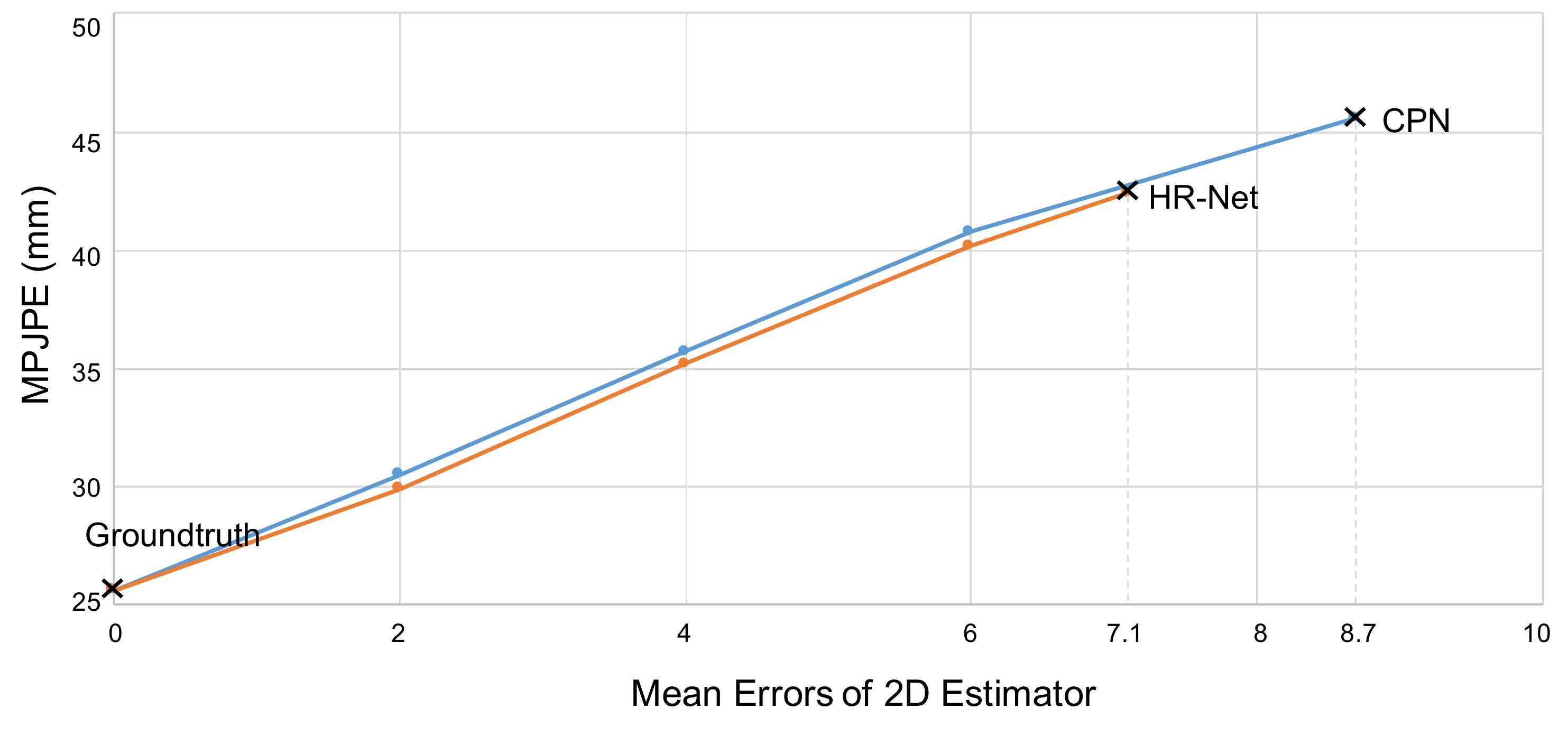}
    \caption{\small
    Relationship between the performace of 3D pose estimation and the accuracy of input 2D poses.
    }
    \label{fig:curve}
\end{figure*}

As shwon in Table 6 of the manuscript,
we achieved a lower MPJPE when using HR-Net~\cite{Sun_2019_CVPR} as the 2D pose estimator than using CPN~\cite{chen2018cascaded}.
To explore the impacts of the 2D pose estimator on the final performance, we combined the predicted 2D pose and the groudtruth by weighted addition for simulating a series of new 2D pose estimators.
UGCN was trained taking as input these synchronized 2D pose.
The results are shwon in the Figure~\ref{fig:curve}.
We can observer a near linear relationship between MPJPE of 3D poses and two-norm errors of 2D poses. Curves from two estimators have very similar tendency.

\clearpage
\section*{Appendix B}
\subsection*{Visual Results}
Visual results of estimated 3D pose by our UGCN are shwon in the Figure~\ref{fig:sample}. More visualized results can be find in the \textbf{supplementary video}, including the following aspects: impacts of motion loss, the comparison with previous works, and the estimation results on noisy 2D poses.

\begin{figure*}[htb]
    \centering
    \includegraphics[width=0.94\linewidth]{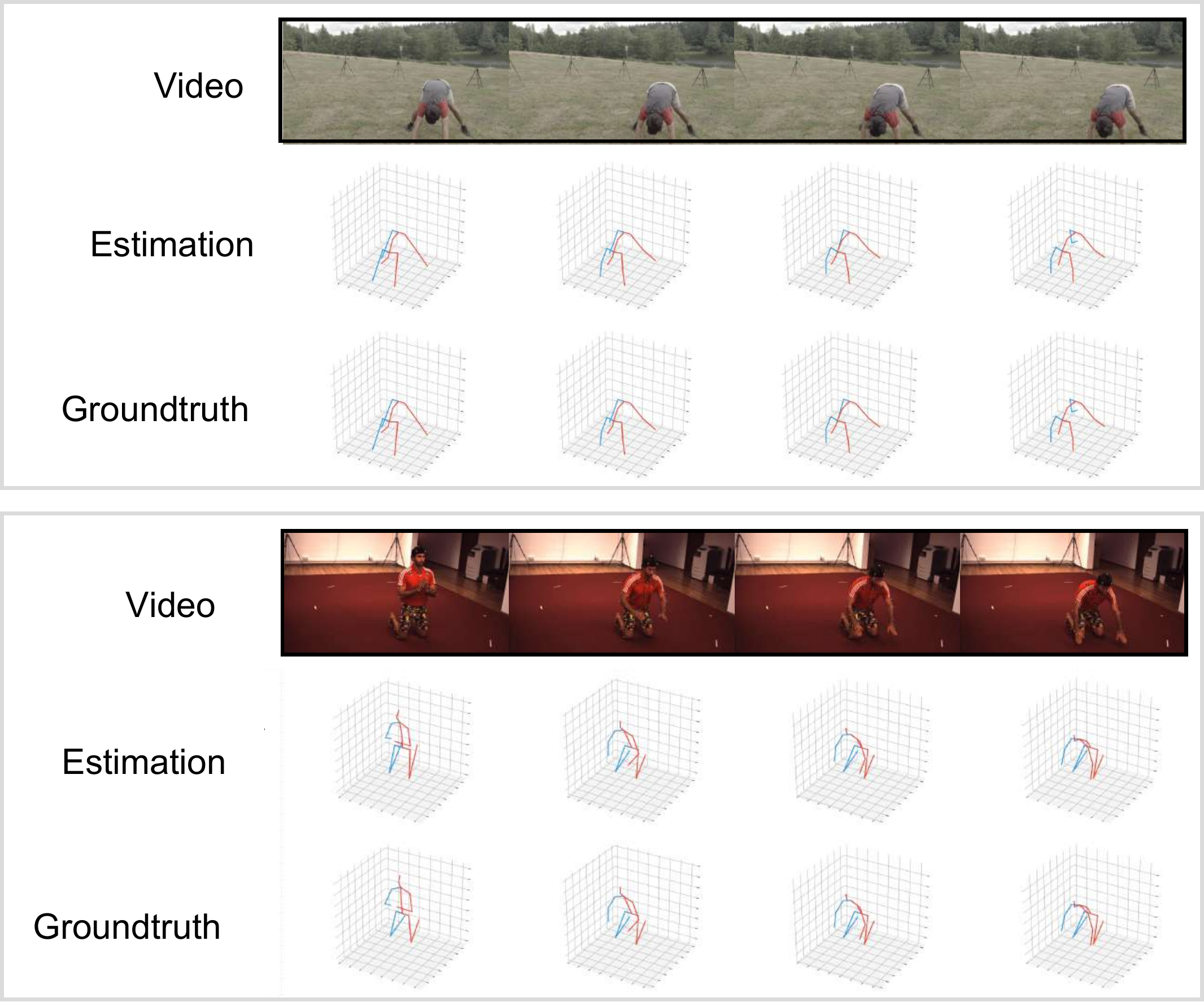}
    \caption{\small
    3D pose sequences estimated by UGCN on two datasets: MPI-INF-3DHP \textbf{(top)} and Human3.6M \textbf{(bottom)}.
    }
    \label{fig:sample}
\end{figure*}

\end{document}